\documentclass[acmsmall]{acmart}

\usepackage{amsmath}
\usepackage{times}
\usepackage{graphicx}
\usepackage{color}
\usepackage{adjustbox}
\usepackage{multirow}

\usepackage{amssymb}
\usepackage{eqparbox}
\usepackage{arydshln}
\usepackage{subcaption}

\AtBeginDocument{%
  \providecommand\BibTeX{{%
    \normalfont B\kern-0.5em{\scshape i\kern-0.25em b}\kern-0.8em\TeX}}}

\setcopyright{acmcopyright}
\copyrightyear{2021}
\acmYear{2021}
\acmDOI{nnn.nnn/nnn.nnn}

\begin{document}

\title{Multi-Domain Spoken Language Understanding Using Domain- and Task-Aware Parameterization}

\author{Libo Qin}
\email{ emails: liboqin@ir.hit.edu.cn}
\affiliation{%
	\institution{Harbin Institute of Technology}
	\department{Research Center for Social Computing and Information Retrieval}
	\city{Harbin}
	\state{Heilongjiang}
	\postcode{150001}
	\country{China}
}
\author{Fuxuan Wei}
\email{ emails: fuxuanwei@ir.hit.edu.cn}
\affiliation{%
	\institution{Harbin Institute of Technology}
	\department{Research Center for Social Computing and Information Retrieval}
	\city{Harbin}
	\state{Heilongjiang}
	\postcode{150001}
	\country{China}
}
\author{Minheng Ni}
\email{ emails: minhengni@ir.hit.edu.cn}
\affiliation{%
	\institution{Harbin Institute of Technology}
	\department{Research Center for Social Computing and Information Retrieval}
	\city{Harbin}
	\state{Heilongjiang}
	\postcode{150001}
	\country{China}
}

\author{Yue Zhang}
\authornote{This is the corresponding author.}
\email{email: yue.zhang@wias.org.cn}
\affiliation{%
	\institution{Westlake University}
	\department{Westlake Institute for Advanced Study}
	\city{Hangzhou}
	\state{Zhejiang}
	\postcode{310024}
	\country{China}
}
\author{Wanxiang Che}
\email{ emails: car@ir.hit.edu.cn}
\affiliation{%
	\institution{Harbin Institute of Technology}
	\department{Research Center for Social Computing and Information Retrieval}
	\city{Harbin}
	\state{Heilongjiang}
	\postcode{150001}
	\country{China}
}
\author{Yangming Li}
\email{ emails: yangmingli@ir.hit.edu.cn}
\affiliation{%
	\institution{Harbin Institute of Technology}
	\department{Research Center for Social Computing and Information Retrieval}
	\city{Harbin}
	\state{Heilongjiang}
	\postcode{150001}
	\country{China}
}
\author{Ting Liu}
\email{ emails: tliu@ir.hit.edu.cn}
\affiliation{%
	\institution{Harbin Institute of Technology}
	\department{Research Center for Social Computing and Information Retrieval}
	\city{Harbin}
	\state{Heilongjiang}
	\postcode{150001}
	\country{China}
}

\renewcommand{\shortauthors}{Qin, et al.}

\begin{abstract}
Spoken language understanding (SLU) has been addressed as a supervised learning problem, where a set of training data is available for each domain. However,  annotating data for a new domain can be both financially costly and non-scalable.
One existing approach solves the problem by conducting multi-domain learning where parameters are shared for joint training across domains, which is \textit{domain-agnostic} and \textit{task-agnostic}. In the paper, we propose to improve the parameterization of this method by using domain-specific and task-specific model parameters for fine-grained knowledge representation and transfer.
Experiments on five domains show that our model is more effective for multi-domain SLU and obtain the best results.
In addition, we show its transferability when adapting to a new domain with little data, outperforming the prior best model by 12.4\%.
Finally, we explore the strong pre-trained model in our framework and find that the contributions from our framework do not fully overlap with contextualized word representations (RoBERTa).
\end{abstract}



\maketitle

\section{Introduction}
\begin{figure}[t]
	\centering
	\includegraphics[width=\textwidth]{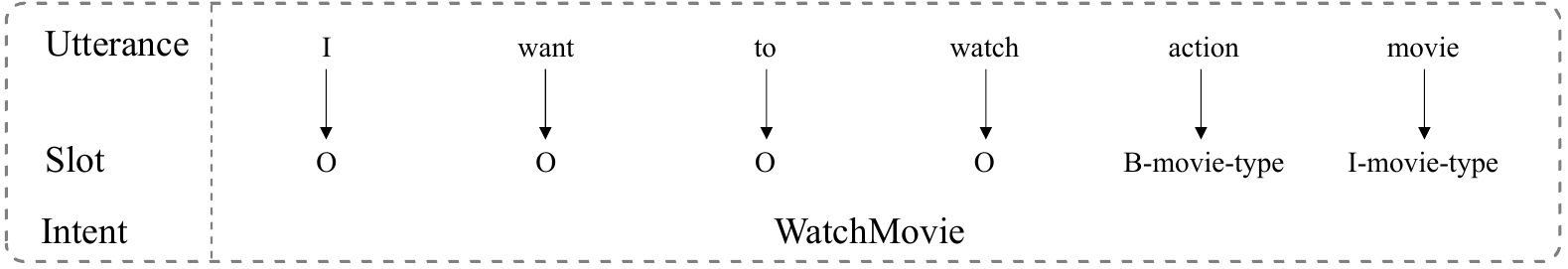}
	\caption{
		An example with intent and slot annotation (BIO format).
	}
	\label{fig:slu}
\end{figure}

Spoken language understanding (SLU) \citep{young2013pomdp} plays an important role in task-oriented dialog systems.
It consists of two typical subtasks, including intent detection and slot filling \citep{tur2011spoken}.
For example in Figure~\ref{fig:slu}, given an input utterance ``\textit{I want to watch action movie}'', the outputs consist of an overall intent class label (i.e., \texttt{WatchMovie}) and a slot label sequence (i.e., \texttt{O}, \texttt{O}, \texttt{O}, \texttt{O}, \texttt{B-movie-type}, \texttt{I-movie-type}).
In particular, the former is a classification task, and the latter can be addressed using sequence labeling.
Since slots highly depend on the intent information, dominant SLU systems in the literature
\citep{goo-etal-2018-slot,li-etal-2018-self,xia-etal-2018-zero,qin-etal-2019-stack,e-etal-2019-novel,liu-etal-2019-cm}
adopt joint models for the two tasks, we follow this line of work, by jointly solving intent detection and slot filling.

\begin{figure}[t]
	\centering
	\includegraphics[width=\textwidth]{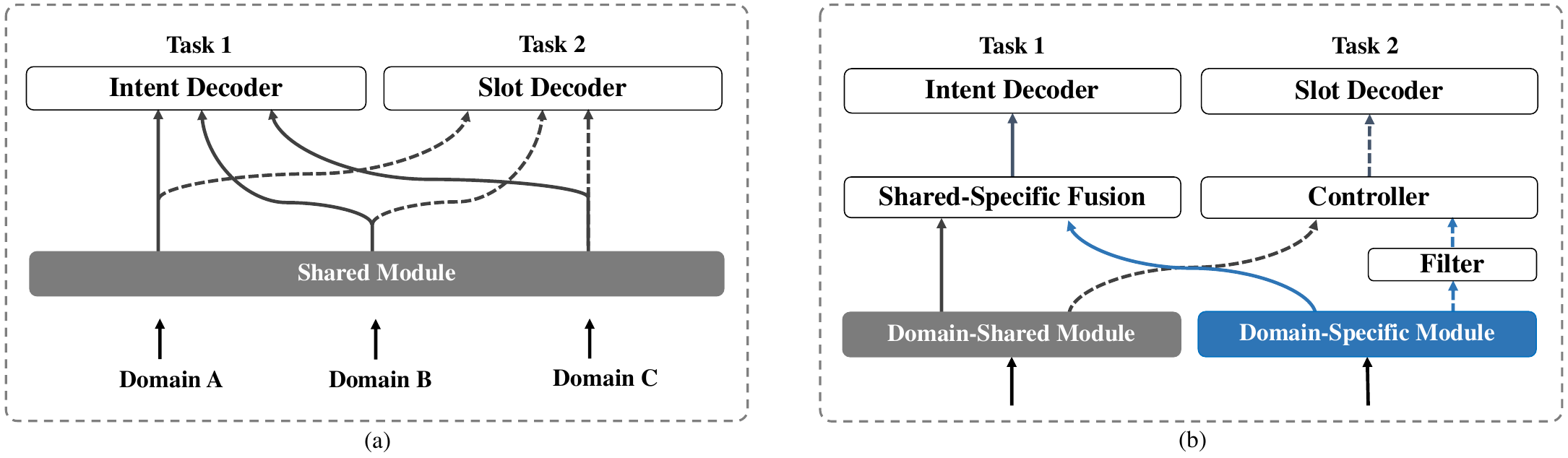}
	\caption{
		Methods for multi-domain spoken language understanding.
		(a) Prior work trains a single model on a mixed dataset.
		(b) Our proposed domain-aware and task-aware model.
		Dash line denotes information flow to slot filling and solid line denotes information flow to intent detection.
		Blue color represents Domain-Specific Module and gray color denotes Domain-Shared Module.
		Better viewed in color.
	}
	\label{fig:contrast}
\end{figure}

Intuitively, there exists a wide range of business domains  (e.g., \textit{watch movie}, \textit{book ticket}) with shared and specific characteristics, and it can be infeasible to train a model for each domain. In practice, a dialogue model should handle multiple domains.
To this end, some existing work has endeavored towards using resources from all domains to train a model \cite{hakkani2016multi,kim2017onenet}. 
As shown in Figure~\ref{fig:contrast}(a), \citet{kim2017onenet} build a model by combining labeled data from different domains for jointly training intent detection and slot filling. 
Their models use the same set of model parameters for representing both cross-domain and cross-task information. 
While this can be useful for feature integration, the method is \textit{domain-agnostic} and \textit{task-agnostic}: 
(1) \textit{domain-agnostic}: One set of shared parameters cannot effectively distinguish the domain-shared and domain-specific features, which limits their performance.
(2) \textit{task-agnostic}: The method represents knowledge for the sentence-level intent detection task and token-level slot filling task equally by using a unified network, and therefore does not offer fine-grained channels for learning task-specific knowledge.
Take the sub-sentence ``\textit{watch action movie}'' for example, the shared and domain-specific knowledge on the words ``\textit{watch}'' and ``\textit{action movie}'' is subtle, because ``\textit{action movie}'' is domain-specific while ``\textit{watch}'' can be shared with other domains \citep{xu2013exploiting}. 
Therefore, domain-specific tokens should be represented with more knowledge from a specific domain, while domain shared tokens should keep shared characteristics across domains.
Unfortunately, solely relying on a unified framework cannot achieve the fine-trained domain knowledge representation and transfer,
which greatly limits its transferability when a new domain with little data is given.

To this end, we propose a domain-aware and task-aware model (i.e., multi-level shared-private framework) for multi-domain joint
intent detection and slot filling, which is shown in Figure~\ref{fig:contrast}(b).
To solve the \textit{domain-agnostic} issue, we first propose to use a standard shared-private framework \citep{liu-etal-2017-adversarial} as a foundation,
which consists of a domain-shared module for representing common knowledge across domains and a domain-specific module for explicitly extracting specific features for each domain.
In addition, utterances from different domains have different sentence syntactic patterns, which helps the model to capture domain-aware features. 
Thus, we explore the domain-aware syntax information and we empirically find modeling syntax information can substantially improve multi-domain SLU.

To address the \textit{task-agnostic} issue, we extend the vanilla shared-private framework to a multi-level structure, achieving the fine-grained domain knowledge transfer : (1) sentence-level domain knowledge transfer is achieved by using a \textit{sentence-level} shared-private architecture for modeling \textit{intents};
(2) token-level domain knowledge transfer is achieved by using a \textit{token-level} shared-private mechanism for modeling \textit{slots}.
Besides, a \textit{slot filter} is applied to each token to selectively decide which tokens receive private representation in addition to a domain-shared representation.
A \textit{slot controller} is further introduced to control the weights between domain-shared and private token representations, 
achieving the fine-grained combination of domain knowledge.

We conduct experiments on two benchmarks, MTOD \citep{schuster-etal-2019-cross-lingual} and ASMixed
\citep{goo-etal-2018-slot,coucke2018snips}, including five different domains in total.
Experiments show that our method achieves state-of-the-art results, with a 91.27\% sentence accuracy on the MTOD dataset, outperforming the prior best result by 1.81\%. 
On the ASMixed dataset, we achieve 84.81\% sentence accuracy, outperforming the prior best result by 3.33\%.
Besides, 
given a new domain with little labeled data, our framework can effectively transfer knowledge
from source training domains, thereby outperforming the existing best model by 12.4\%. 

Finally, Pre-trained models (PLMs) have achieved surprising results across almost all NLP tasks.  A natural question is raised whether our framework can still obtain improvement over the pre-trained model.
To answer this question,
we explore the pre-trained model (RoBERTa) \citep{liu2019roberta} in our framework and find that our framework works orthogonally with pre-trained model.

In summary, the contributions of our work can be concluded as follows: 

\begin{itemize}
	\item[$\bullet$] We propose a domain-specific and task-specific parameterize method (i.e., multi-level shared-private framework) for multi-domain SLU, the structure of which is domain-aware and task-aware, which greatly improves the transferability across domains.
	\item[$\bullet$] We propose a \textit{token-level} shared-private mechanism, which enables the model to achieve a fine-grained domain knowledge fusion for slot filling. 
	To the best of our knowledge, this is the first attempt to consider the fine-grained knowledge transfer for multi-domain SLU.
	\item[$\bullet$]  Experiments on two benchmarks show that our framework obtains substantial improvement over existing multi-domain SLU methods and achieves state-of-the-art performance.
	\item[$\bullet$] We explore and analyze the effect of incorporating pre-trained model (RoBERTa) in multi-domain SLU tasks and empirically shows that  the
	contributions from our framework 
	do not
	fully overlap with contextualized word representations.
\end{itemize}

\section{Task Defination}
\paragraph{Intent Detection:}
Given input utterance $X$ = $(x_{1},\dots,x_{n})$ ($n$ denotes the length of $X$), intent detection (ID) can be considered as a sentence classification task to decide the intent label $o^{I}$, which is formulated as:
\begin{equation}
o^{I} =  \texttt{Intent-Detection} (X).
\end{equation}
\paragraph{Slot Filling:}
Slot filling (SF) can be seen as a sequence labeling task to produce a sequence slots $o^{S}$ = $(o^{S}_{1},\dots,o^{S}_{n})$, which can be written as:
\begin{equation}
o^{S} = \texttt{Slot-Filling} (X).
\end{equation}
\paragraph{Joint Model:}
Joint model denotes that a joint model predicts the slots sequence and intent simultaneously, which has the advantage of capturing shared knowledge across related tasks, using:
\begin{equation}
(o^{I}, o^S)  =  \texttt{Joint-Model} (X).
\end{equation}
\paragraph{Multi-Domain Learning}
Suppose that there is a set of domains $D = \left\{d_1, d_2, ..., d_{|D|}\right\}$ and a dataset with $m$ data instances $T = \left\{t_1, t_2, ..., t_m\right\}$.
For each $t$ in $T$, we have $t = (X, \textbf{o}^{S}, {o}^{I}, d)$, where $X$ represents utterance, $\textbf{o}^{S}$ represents target slots, ${o}^{I}$ represents target intent and $d$ represents the domain of this data, respectively.
The goal is to train a joint model on multiple source domains, which can be used for each domain.

\section{Approach}
\begin{figure*}[t]
	\centering
	\includegraphics[scale=0.5]{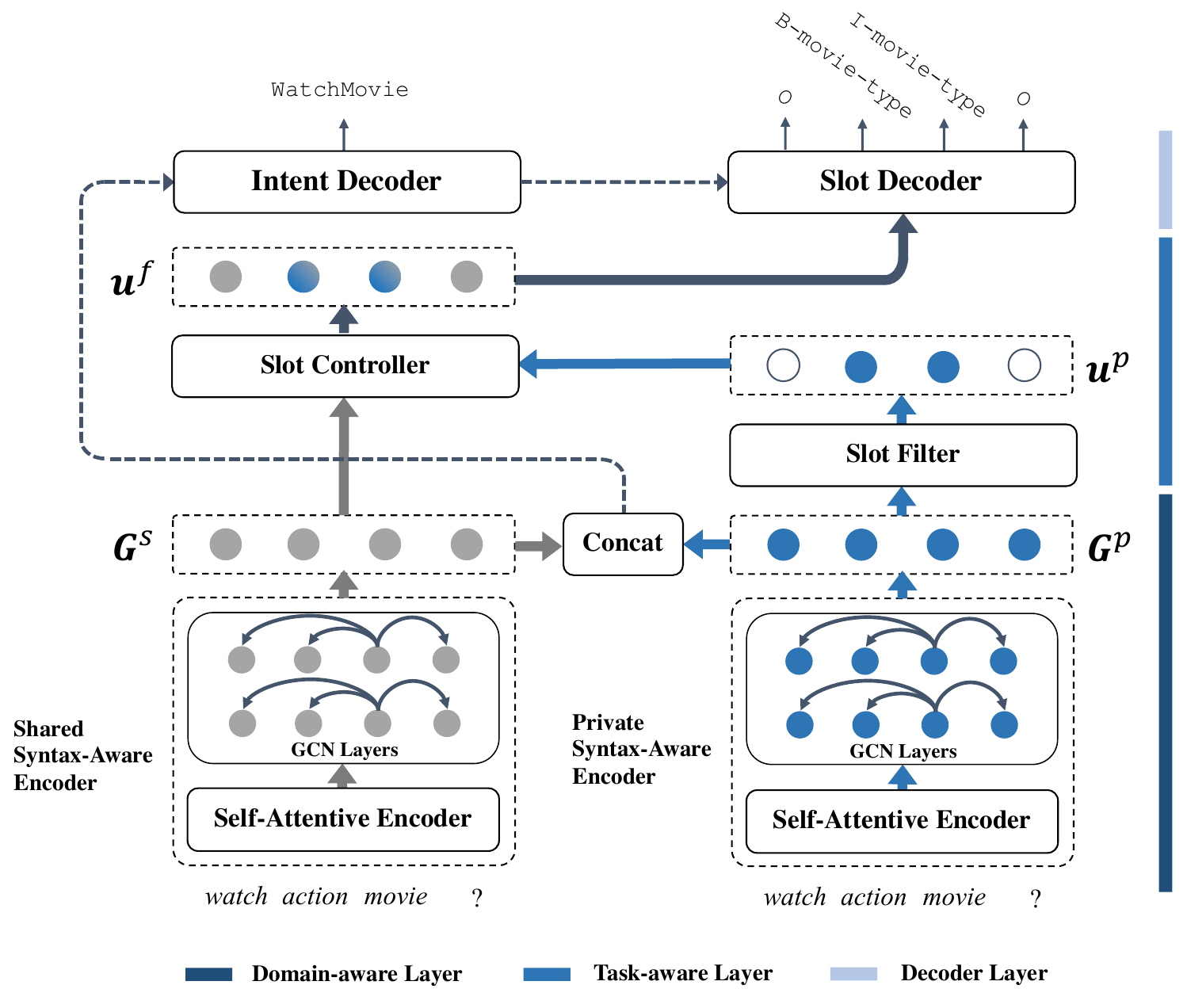}
	\caption{
		Overview of our proposed framework. 
		It consists of a shared-private syntax-aware encoder, a slot filter, a slot controller and two decoders.
		The gray color represents general features across all domains and the blue color denotes domain-specific features. 
		For simplicity, we only draw the shared encoder and the encoder of the $d^{th}$ domain in the figure.
		Better viewed in color.
	}
	\label{fig:framework}
\end{figure*}
The overall structure of our multi-level shared-private framework is shown in Figure~\ref{fig:framework}. 
First, the model consists of a \textit{shared-private syntactic encoder}, which is used for generating domain-shared and domain-specific features.
Second, the sentence-level
shared and private features are combined for intent detection directly.
Third, a token-level shared-private framework is used on top of the sentence-level representations, which includes a  two-stage decoder, and for making fine-grained knowledge transfer for slot filling.

\subsection{Shared-private Syntactic Encoder}
As shown in Figure~\ref{fig:framework}, a shared syntax-aware encoder is used to capture domain-shared features. 
Each instance passed into the shared encoder and its corresponding private encoder to obtain the representation.

\paragraph{Self-Attentive Sentence Representation} Following \citet{qin-etal-2019-stack},  we first use a basic self-attentive encoder to obtain self-attentive representation, whic{\tiny }h includes a bidirectional LSTM (BiLSTM) \citep{hochreiter1997long} to obtain the temporal information within words and a self-attention mechanism to capture the contextual information.
Given an $n$-word sentence ${X}$ = (${{x}}_{1}, {{x}}_{2},.., {{x}}_{n}$), we first use the BiLSTM to read it 
forwardly from ${x}_{1}$ to ${x}_{n}$ and backwardly from ${x}_{n}$ to ${x}_{1}$ to produce a series of context-sensitive hidden states $\boldsymbol{H} = \{\boldsymbol{h}_{1}, \boldsymbol{h}_2, \ldots, \boldsymbol{h}_{n}\}$, which can be denoted as:
\begin{equation}
\begin{aligned}
& \overrightarrow{\boldsymbol{h}_{i}} = 	\overrightarrow{\operatorname{LSTM}}(\phi^{\text{emb}} (x_{i}), 	\overrightarrow{\boldsymbol{h}_{i-1}}), i \in [1, n] \,, \\
&\overleftarrow{\boldsymbol{h}_{i}} = 	\overleftarrow{\operatorname{LSTM}}(\phi^{\text{emb}} (x_{i}), 	\overleftarrow{\boldsymbol{h}_{i+1}}), i \in [n, 1] \,, \\
& \boldsymbol{ h}_{i} = [\overrightarrow{\boldsymbol{h}^{}_i},\overleftarrow{\boldsymbol{h}^{}_i}],
\end{aligned}
\label{eq:lstm}
\end{equation}
where $\phi^{emb}(\cdot)$ denotes the embedding function.

Self-attention is a very effective method of leveraging context-aware features over variable-length sequences for natural language processing tasks \cite{zhong-etal-2018-global}. Therefore, we also apply self-attention over word embedding to capture context-aware features. 
We adopt a Transformer encoder \cite{NIPS2017_7181}, which
maps the matrix of input vectors $\boldsymbol{X}$ = 
$\{\phi^{emb}(x_{1}), \ldots, \phi^{emb}(x_{n})\}$  $\in$ $\mathbb{R}^{n\times d}$ 
($\phi^{emb}$ represents embedding mapping matrix)
to queries (${ \boldsymbol{Q}}$), keys ($ \boldsymbol{K}$) and values 
($ \boldsymbol{V}$) matrices by using different linear projections
and output 
${ \boldsymbol{C}}$ $\in$ $\mathbb{R}^{T\times d}$  is a weighted sum of values:
\begin{equation}
{\boldsymbol{C}} = \operatorname { softmax } \left( \frac {  \boldsymbol{Q}  \boldsymbol{K}^\top } { \sqrt { d _ { k } } } \right) \boldsymbol{V},
\end{equation}
where $d_{k}$ denotes the dimension of keys.
We concatenate these two representations as 
the self-attentive encoding representation:
\begin{equation}
\boldsymbol{ E} =\boldsymbol{ H } \oplus\boldsymbol{C},
\end{equation}
where {$\boldsymbol{E}$} = (${\boldsymbol{e}}_{1}, {\boldsymbol{e}}_{2},.., {\boldsymbol{e}}_{n}$) $\in$ $\mathbb{R}^{n\times 2d}$ and $\oplus$ is concatenation operation. 

\paragraph{Graph Convolution over Dependency Trees}
Syntax information is an important source of features across domains.
We use a GCN \citep{kipf2016semi} over the dependency tree of a sentence\footnote{We use Stanford CoreNLP \citep{manning2014stanford} to generate the dependency tree. } to exploit syntactic information.
Given a graph with $k$ nodes, an adjacency matrix
$\boldsymbol{A}\in\mathbb{R}^{k\times k}$ is used to represent the graph, where $A_{ij} = 1$ if there is an edge going from node $i$ to node $j$. 
We denote the $l$-th layer output for node $i$ as $\boldsymbol{g}_i^{(l)}$, where $\boldsymbol{g}_i^{(0)}$ represents the initial state of node $i$. 

Following \citet{zhang-etal-2018-graph,zhang-etal-2019-aspect}, we set $\boldsymbol{G}^{(0)}$ = $\boldsymbol{E}$. 
Given the dependency tree of the input sentence, the graph convolution operated on the node representation can be written as:
\begin{eqnarray}
\boldsymbol{g}_i^{(l)} = \sigma\big( \sum_{j=1}^n \tilde{A}_{ij} \boldsymbol{W}^{(l)}\boldsymbol{g}_j^{(l-1)} + \boldsymbol{b}^{(l)} \big),\label{eqn:gcn_final}
\end{eqnarray}
where the layer $l\in [1,2,\cdots,L]$, ${\boldsymbol{\tilde{A}}} = \boldsymbol{A} + \boldsymbol{I}$ and $\boldsymbol{I}$ is a $n\times n$ identity matrix to consider information itself, $ \boldsymbol{W}^{(l)}$ is a linear transformation, $ \boldsymbol{b}^{(l)}$ is a bias term and $\sigma$ is a nonlinear function. $\boldsymbol{g}_i^{(L)}$ is the final state of node $i$.

For shared-private modeling, we allocate a set of parameters $\Theta$ shared across all domains, and a set of private parameter $\Theta^p_d$ for each domain, d $\in$ $\{1, ..., |D|\}$. Thus the total set of model parameters is $\Theta \bigcup \Theta^p_1, \dots, \Theta^p_D$ .
For testing, given an input utterance, we use $\Theta^s$ to calculate a shared representation $\boldsymbol{G}^s$, and $\Theta^p_d$ that corresponds to the input domain to calculate a private representation $\boldsymbol{G}^p$.

\subsection{Domain-aware Sentence-level Transfer for Intent Detection}
We use a standard sentence-level shared-private structure  \citep{yang2016multi, liu-etal-2017-adversarial} over the input utterance for knowledge transfer concerning intent detection.

\paragraph{Domain Shared-Private Feature Fusion}
After obtaining the domain-shared and domain-specific encoding representation $\boldsymbol{G}^{{s}}$, $\boldsymbol{G}^{{p}}$, we use self-attention \citep{P18-1135,goo-etal-2018-slot} to aggregate relevant context representation for intent detection:
\begin{eqnarray}
\label{eq:selfattnscore}
\boldsymbol{a}^{{s}} &=& \boldsymbol{W}^{{s}} \boldsymbol{G}^{{s}} + \boldsymbol{b}^{{s}}, \\
\boldsymbol{p}^{{s}} &=& \operatorname { softmax } \left( \boldsymbol{a}^{{s}} \right).
\end{eqnarray}

The shared context representation $\boldsymbol{c}^{{s}}$ is computed as the sum of each element $\boldsymbol{g}_i^{{s}}$, weighted by the corresponding normalized self-attention score $p^{{s}}_i$:
\begin{eqnarray}
\label{eq:selfattncontext}
\boldsymbol{c}^{{s}} &=& \sum_i p^{{s}}_i \boldsymbol{g}^{{s}}_i. 
\end{eqnarray}

We similarly compute the local self-attention context $\boldsymbol{c}^{{p}}$.

$\boldsymbol{c}^{{s}}$ and $\boldsymbol{c}^{{p}}$ are combined by concatenation:
\begin{eqnarray}
\boldsymbol{c}^{m} = [\boldsymbol{c}^{{s}}, \boldsymbol{c}^{{p}}],
\end{eqnarray}
deriving a mixed syntax-aware encoding feature $\boldsymbol{c}^{m}$.
\paragraph{Intent Prediction}
The mixed encoding vector $\boldsymbol{c}^{m}$ 
is used as input for intent detection:
\begin{eqnarray}
{\boldsymbol{y}} ^ { I } &=& \operatorname { softmax } \left( {\boldsymbol{W}} _ { h } ^ { I }{{\boldsymbol{c}^{m}}}  \right), \\
{o}  ^ { I } &=& \operatorname { argmax } ({\boldsymbol{y}}  ^ { I }),
\end{eqnarray}
where ${\boldsymbol{y}} ^ { I }$ is the output intent distribution; ${{o}^{I}} $ represents the intent label and ${\boldsymbol{W}}_{h}^{I}$ are trainable parameters of the model.

\subsection{Task-aware Token-Level Transfer for Slot Filling}
A token-level shared-private network is used to model the task-aware token-level transfer for slot filling.
Specially, we use a two-stage decoder to consider task characteristics for slot filling, building on top of the shared-private encoder as shown in Figure~\ref{fig:framework}.
The first stage uses a \textit{filter} to mask out those domain-general tokens, which do not need domain-specific features\footnote{We treat non-slot labels as domain-general. (e.g. slots that are tagged \texttt{O}).}, so as to allow our model to focus on knowledge transfer for more inferable tokens.
The second stage makes use of a \textit{controller} module to achieve fine-grained knowledge transfer by automatically calculating the weights for domain-shared and domain-specific features at the token-level.

\paragraph{Slot Filter}
We adopt a simple feedforward network as our filter.
$\boldsymbol{G}^{{s}}$ and $\boldsymbol{G}^{{p}}$ are concated as input to the filter module:
\begin{eqnarray}
{\boldsymbol{F}} &=& \operatorname { sigmoid } \left( {\boldsymbol{W}} _ {f}{{[\boldsymbol{G}^{{s}}; \boldsymbol{G}^{{p}}]} +\boldsymbol{b}_f}  \right), 
\end{eqnarray}
where  ${\boldsymbol{W}}_{f}$ are trainable parameters, and we define the label $\boldsymbol{F}$ = $(f_{1},...,f_{n})$ as the probability of domain-general tokens. Correspondingly, $(1- \boldsymbol{F})$ = $(1-f_{1}),...,(1- f_{n})$ is the probability of domain-specific tokens.

We use the output of the filter module on ${\boldsymbol{G}^{{p}}}$, where
\begin{eqnarray}
\boldsymbol{u}^{{p}}_{i} & = & (1-f_i) \cdot \boldsymbol{g}^{{p}}_i.
\end{eqnarray}

The resulting vectors $\boldsymbol{U^{p}} =\{\boldsymbol{ u}^{{p}}_{1},\dots,\boldsymbol{ u}^{{p}}_{n}\}$ represent domain-specific features.

Given $\boldsymbol{ U^{p}}$, we use a controller to generate weights on domain-shared and domain-specific features at the token-level, making a fine-grained fusion between the domain-shared and private features at the token-level.
\paragraph{Slot Controller}
We concatenate $\boldsymbol{G}^{{s}}$, $\boldsymbol{G}^{{p}}$ and use a simple feedforward network to calculate weights at each token, which can be written as follows:
\begin{eqnarray}
\label{eq:selfattnscore}
{\boldsymbol{P}}  &=& \operatorname { sigmoid } \left( {{\boldsymbol{W}_c}}[\boldsymbol{G}^{{s}}; \boldsymbol{G}^{{p}}] +\boldsymbol{b}_c\right).
\end{eqnarray}

The weights $\boldsymbol{ P}$ = $\{p_{1}, \dots, p_{n}\}$ produced by the controller module are used to fuse domain-shared and domain-specific features
\begin{eqnarray}
\boldsymbol{u}^{f}_{i} & = & {p} _ { i } \cdot \boldsymbol{u}^{{p}}_{i}   + (1 - {p} _ { i }) \cdot \boldsymbol{g}^{{s}}_i,
\end{eqnarray}
where $\boldsymbol{u}^{f}_{i}$ is the fused representation at $i^{th}$ token. 
\paragraph{Slot Prediction}
We use a unidirectional LSTM as the slot-filling decoder. 
Following \citet{li-etal-2018-self} and \citet{qin-etal-2019-stack}, we adopt intent information to guide the slot prediction.
At the $i^{th}$ decoding step, the decoder state ${\boldsymbol{h}}_{i}^{S}$ 
can be formalized as:
\begin{equation}
{\boldsymbol{h}} _ { i }^{S} = \operatorname{LSTM} \left( {\boldsymbol{h}} _ { i - 1 }^{S} , {\boldsymbol{y}} _ { i - 1 }^{S} ,  {\boldsymbol{y}}  ^ { I } \oplus{{\boldsymbol{u}^{f}_{i}}}\right),
\end{equation}
where ${\boldsymbol{h}}_{i-1}^{S}$ is the previous decoder state;  
${\boldsymbol{y}}_{i-1}^{S}$ is the previous emitted slot label distribution and $\boldsymbol{y
}^{I}$ is embedding of intent.

Finally, ${\boldsymbol{h}}_{i}^
{S}$ is used for slot prediction:
\begin{eqnarray}
{\boldsymbol{y}} _ { i } ^ { S } &=& \operatorname { softmax } \left( {\boldsymbol{W}} _ { h } ^ { S } {\boldsymbol{h}}_ { i }^{S}\right),\\
{o} _ { i } ^ { S } &=& \operatorname { argmax } ({\boldsymbol{y}} _ { i } ^ { S }),
\end{eqnarray}
where ${o} _ { i } ^ { S }$ is the slot label of the $i^{th}$ word in the utterance.

\subsection{Joint Training}

We adopt a joint model to consider the two tasks and update parameters in a joint optimization.
A cross-entropy loss is used for intent detection:
\begin{equation}
\mathcal { L } _ { 1 } \triangleq - \sum _ { j = 1 } ^ { m }   {\hat { {\boldsymbol{y}} }  ^ { j, I } } \log \left( {\boldsymbol{y}}  ^ { j, I } \right).
\end{equation}
Similarly, the slot filling objective is: 
\begin{equation}
\mathcal { L } _ { 2 } \triangleq - \sum _ { j = 1 } ^ { m } \sum _ { i = 1 } ^ { n_j } {  {\hat { {\boldsymbol{y}}} _ { i } ^ { j,S } } \log \left( {\boldsymbol{y}} _ { i } ^ {j, S } \right)},
\end{equation}
where ${\hat { {\boldsymbol{y}} } _ { j } ^ { I } }$ and 
$ {\hat { {\boldsymbol{y}}} _ { i } ^ { S } }$ are the gold intent label and gold slot label, respectively;  $m$ is the number of training data and $n_j$ is the number of tokens in $j^{th}$ data.

In addition, to further strengthen the filter, we add an auxiliary loss to train the Filter as a classification task.
The loss function can be denoted as:
\begin{equation}
\begin{split}
\mathcal { L } _ { 3 } \triangleq - \sum _ { j = 1 } ^ { m } \sum _ { i = 1 } ^ { n_j } {  { \hat{{y}} } _ { i } ^ {j, F } \log \left( {{f}} _ { i } ^ {j } \right)
	+ (1 - { \hat{{y}} } _ { i } ^ {j, F }) \log \left(1- {{f}} _ { i } ^ {j } \right)} ,
\end{split}
\end{equation}
where ${\hat{y}} _ { i } ^ {j, F }$ is the gold representation of Filter.
The final joint objective is formulated as:
\begin{equation}
\mathcal { L } _ { \theta } = \alpha_{1} \mathcal{ L }_{1} +\alpha_{2}  \mathcal{ L }_{2}
+ \alpha_{3} \mathcal{ L }_{3},
\end{equation}
where $\alpha_{1}$, $\alpha_{2}$ and $\alpha_{3}$ are hyper-parameters.

\begin{table}[t]
	\begin{center}
		\begin{adjustbox}{width=0.6\textwidth}
			\begin{tabular}{l|l|l|l|l}
				\hline 
				\textbf{Dataset} &\textbf{ Domains}&\textbf{Train} & \textbf{Dev} & \textbf{Test} \\ 
				\hline
				\multirow{1}*{MTOD} 
				&Reminder, Weather, Alarm& 30,521 & 4,181 & 8,621 \\
				
				\hline
				\multirow{1}*{ASMixed} 
				&ATIS, SNIPS& 17,562 & 1,200 & 1,593 \\
				\hline
			\end{tabular}
		\end{adjustbox}
		\caption{Statistics of datasets.}\label{tab:datasets}
	\end{center}
\end{table}
\section{Experiments}
\subsection{Datasets}
We conduct experiments on the benchmark MTOD \citep{schuster-etal-2019-cross-lingual}\footnote{The reason why we do not adopt multiwoz is that multiwoz is mainly proposed to evaluate the dialog state tracking task rather than the spoken language understanding, which makes it hard for us to directly use it as a benchmark for evaluating the multi-domain SLU task. }.
The dataset contains three domains including Alarm, Reminder, Weather domain.
We follow the same format and partition as in \citep{schuster-etal-2019-cross-lingual}.
To verify the generalization of the proposed model, 
we construct another multi-domain SLU dataset (ASMixed) by mixing the ATIS  \citep{hemphill1990atis} and SNIPS \citep{coucke2018snips} dataset and keeping the train/dev/test partition unchanged.
The detailed statistics of the two datasets are shown in Table~\ref{tab:datasets}.

\subsection{Experimental Settings} \label{sec:setting}
The dimensionalities of the embeddings are $64$ and of the LSTM hidden units are $256$.
The dropout ratio is 0.4 and the batch size is 16. The learning rate is 0.001.
The GCN layer number is 3 for MTOD and 2 for ASMixed.
In the framework, we use Adam \citep{kingma-ba:2014:ICLR} to optimize the model parameters and 
adopt the suggested hyper-parameters.

All experiments are conducted
 using GeForce RTX 2080Ti GPU. The epoch number is 100 for two datasets and we do not adopt early stopping strategy.
 For all experiments, we pick the model which works best on development set, and then evaluate it on test set.

Similar to \texttt{One-Net} \citep{kim2017onenet}, we assume that the input is one utterance, without the need to know which domain it comes from. 
During the test period, we adopt a syntax-aware encoder, which is the same as the shared syntax-aware encoder shown in Figure~\ref{fig:framework}, to directly predict the domain of a given utterance.
The result is 99.9\% in MTOD dataset and 99.7\% in ASMixed dataset.
The domain classification is very high due to the explicit features in spoken language utterances,
which is consistent with the observation of \citet{gupta2018efficient}.
\subsection{Baselines} \label{sec:baselines}
We compare our model with several existing state-of-the-art multi-domain SLU baselines including:\\
1) \texttt{Shared-LSTM}
\citet{hakkani2016multi} trained a single model for intent detection and slot filling using data from all the domains, which has advantage of incorporating domain-shared knowledge..\\
2) \texttt{Separated-LSTM}. \citet{hakkani2016multi} proposed a
single-domain joint model for slot filling and intent detection, which can capture domain-specific features for each domain..\\
3) \texttt{Multi-Domain Adv}. \citet{liu2017multi} applied an adversarial training method for
slot filling.
For a fair comparison, we add an intent detection module and train the two tasks jointly.\\
4)  \texttt{One-Net}. \citet{kim2017onenet} used data-combined for joint slot filling and intent detection, which is another parameter-shared method to incorporate domain-shared features.. \\
5) \texttt{Locale-agnostic-Universal}.
\citet{lee-etal-2019-locale}
proposed a locale-agnostic universal
domain classification model based on multi-task learning. \\
6) \texttt{Coach}.
\citet{liu2020coach} proposed a coarse-to-ﬁne approach (Coach) for cross-domain slot ﬁlling. Besides, slot descriptions are used in the fine stage to help recognize unseen slots, and template regularization is applied to further improve the slot ﬁlling performance of similar or the same slot types.
This approach achieves the state-of-the-art performance.
For a fair comparison, we add the intent detection upon coach model for joint SLU task.

\begin{table}[t]
	\centering
	\begin{adjustbox}{width=\textwidth}
		\begin{tabular}{l||cccccc||ccccc}
			\hline
			& \multicolumn{6}{c}{MTOD}  & \multicolumn{5}{c}{ASMixed}\\
			\hline \hline
			\textbf{Model} & \textbf{Overall Exact}  & \textbf{Slot}  & \textbf{Intent}  &
			\begin{tabular}[c]{@{}c@{}}\textbf{Reminder} \\ \textbf{Exact}\end{tabular} & \begin{tabular}[c]{@{}c@{}}\textbf{Alarm} \\\textbf{Exact}\end{tabular} 
			& \begin{tabular}[c]{@{}c@{}}\textbf{Weather} \\\textbf{Exact}\end{tabular} &\textbf{Overall Exact} & \textbf{Slot}  & \textbf{Intent} &
			\begin{tabular}[c]{@{}c@{}}\textbf{ATIS} \\\textbf{Exact}\end{tabular} & 
			\begin{tabular}[c]{@{}c@{}}\textbf{SNIPS} \\\textbf{Exact}\end{tabular} \\
			\hline
			Shared-LSTM \citep{hakkani2016multi} & 88.71 & 94.87 & 98.70 & 82.06 & 90.19 & 90.99 & 76.71 & 92.55 & 94.41 & 81.69 & 70.41\\
			Separated-LSTM \citep{hakkani2016multi} & 89.73 & 94.89 & 99.01 & 84.59 & 89.81 & 92.18 & 79.53 & 92.94 & 94.79 & 80.96 & 77.71\\
			Multi-Domain adv \citep{liu2017multi} & 88.82 & 94.41 & 98.87 & 82.09 & 88.86 & 92.05 & 79.47 & 91.80 & 96.48 & 82.75 & 75.29 \\
			One-Net \citep{kim2017onenet} & 89.36 & 95.25 & 98.56 & 83.27 & 90.15 & 91.83 & 78.28 & 93.38 & 93.72 & 81.85 & 73.80\\
			
			Locale-agnostic-Universal \citep{lee-etal-2019-locale} & 88.54 & 94.16 & 99.12 & 81.63 & 89.58 & 91.21 & 79.35 & 92.10 & 96.48 & 82.19 & 75.71\\
			
			Coach \citep{liu2020coach} & 89.46 & 95.13 & 98.38 & 83.02 & 90.69 & 91.78 & 81.48 & 92.87 & 96.86 & 82.49 & 80.20\\
			\hline
			Ours Framework & \textbf{91.27*} & \textbf{95.69*} & \textbf{99.20*} & \textbf{85.62*} & \textbf{92.37*} & \textbf{93.29*} & \textbf{84.81*} & \textbf{94.30*} & \textbf{97.30*} & \textbf{86.53*} & \textbf{82.62*}\\
			\hline
		\end{tabular}
	\end{adjustbox}

	\caption{Main Results (Overall Exact, Slot and Intent denote the corresponding metrics on whole datasets and domain exact represents the exact accuracy on each domain separately). The numbers with * indicate that the improvement of our framework overall baselines is statistically significant with $p < 0.05$ under t-test.} \label{tab:results}
	\vspace{-0.3cm}
\end{table}

\subsection{Overall Results}
Following prior work \citep{goo-etal-2018-slot, qin-etal-2019-stack}, we evaluate the performance of slot filling using 
F1 score, intent prediction using accuracy 
and sentence-level semantic frame parsing 
using the exact accuracy, measuring the ratio of sentences for which both intent and slot are predicted correctly in a sentence.

Table~\ref{tab:results} shows the results of the proposed models
on two datasets.
We can observe that:
\begin{itemize}
	\item[$\bullet$] \texttt{Locale-agnostic-Universal} achieves the best performances on intent detection among all baselines, which indicates that explicitly modeling domain-shared and domain-specific is more effective than implicitly capturing shared knowledge with sharing parameters.

	\item[$\bullet$] Our model
	significantly outperforms all the baselines by a 
	large margin and achieves state-of-the-art 
	performance.
	In particular, on the ASMixed dataset, compared 
	with the best prior joint work \texttt{Coach},
	we achieve 1.43\% improvement on slot filling task. On the MTOD dataset, the same trend has been witnessed.
	This indicates the effectiveness of our task-aware token-level shared-private framework, which can effectively transfer fine-grained knowledge for each domain.
	
	\item[$\bullet$] Our framework gains the largest improvements on overall exact. We attribute this to the fact that our proposed domain-aware and task-aware framework can better help transfer domain knowledge between the intent and slots and hence improve the SLU performance.

\end{itemize}

\begin{figure*}[t]
	\centering
	\begin{adjustbox}{width=1.0\textwidth}
		\begin{subfigure}[t]{0.5\linewidth}
			\includegraphics[width=1.0\linewidth]{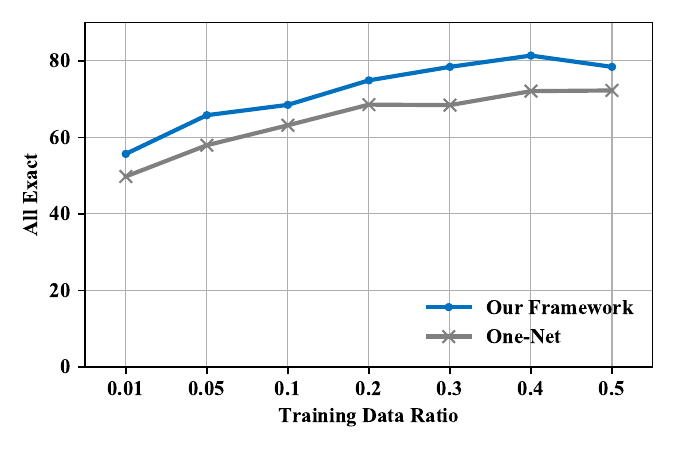}
			\caption{Overall Exact (Acc).}
			\label{fig:myfig3}
		\end{subfigure}
		\quad
		\begin{subfigure}[t]{0.5\linewidth}
			\includegraphics[width=1.0\linewidth]{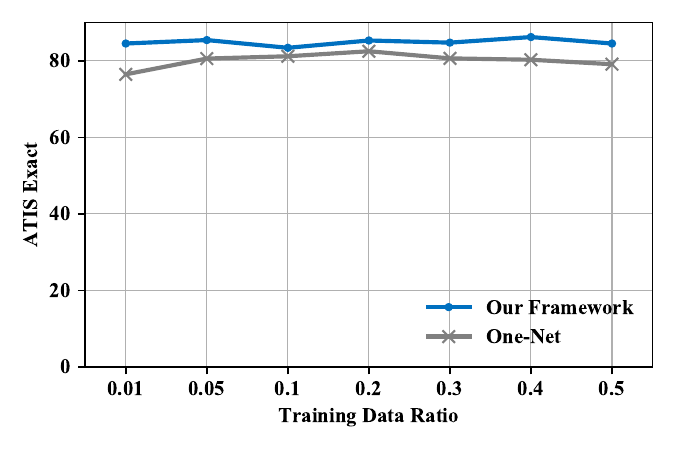}
			\caption{ATIS Exact (Acc).}
			\label{fig:myfig4}
		\end{subfigure}
		\quad
		\begin{subfigure}[t]{0.5\linewidth}
			\includegraphics[width=1.0\linewidth]{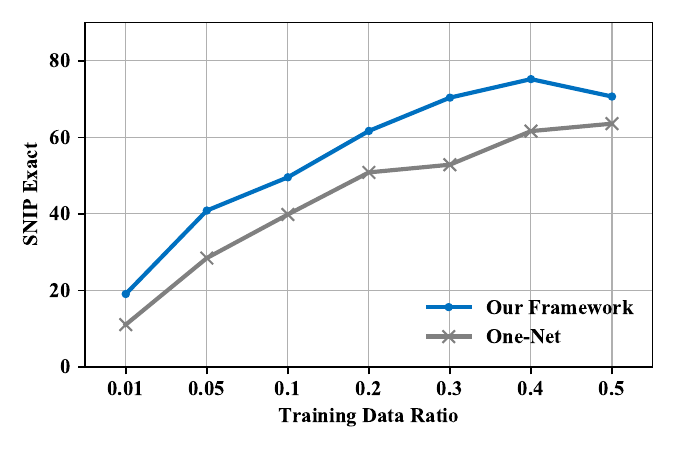}
			\caption{SNIPS Exact (Acc)}
			\label{fig:myfig5}
		\end{subfigure}
	\end{adjustbox}
	\caption{Performance (Exact Acc) of domain adaption on different subsets of the original training data on the ASMixed dataset.}
	\label{fig:few-shot}
\end{figure*}

\subsection{Analysis}
More thorough studies and analysis are conducted  on ASMixed in this section, trying to answer the following questions:
\begin{enumerate}
	\item Does the domain- and task-aware multi-level shared-private module benefit multi-domain SLU?
	\item Does the domain-aware sentence-level shared-private module benefit multi-domain SLU?
	\item Does the task-aware two-stage decoder successfully transfer fine-grained knowledge for token-level slot filling across all domains? 
	\item Can domain-aware syntactic information better generalize across domains in SLU tasks?
	\item Can our framework effectively transfer knowledge for a new domain with little labeled data? 
	\item Does our framework successfully capture domain-shared and domain-specific features?
	\item Does our framework still works upon pre-trained model?
\end{enumerate}

\subsubsection{Effectiveness of Domain- and Task-Aware Multi-level Shared-private Framework}\label{sec:domain}
In order to verify the effectiveness of the proposed domain- and task-aware multi-level shared-private architecture, we
conduct a set of experiments where the private syntax-aware encoder is removed. This means that the ablated framework cannot access to domain-specific features.
To see the effectiveness of this module fairly, we keep the two-stage module unchanged and use a shared encoder to replace the original private encoder.
The results are shown in the 
\textit{w/o Multi-level Shared-private Architecture} row in Table \ref{tab:ablation}.
We observe that the ATIS Exact acc drops by 1.57\% and SNIPS Exact acc drops by 2.28\% when the shared-private framework is removed.
In addition, the whole exact acc drops by 1.88\%, which shows that domain-specific feature extracted by the private module is important for multi-domain SLU, including both the sentence-level intent detection subtask and token-level slot filling sub task.
\begin{table}[ht]
	\centering
	
	\begin{adjustbox}{width=1.0\textwidth}
		\begin{tabular}{l|ccc|cc}
			\hline
			\textbf{Model} & \begin{tabular}[c]{@{}c@{}}\textbf{Overall} \textbf{Exact}\end{tabular}  & \textbf{Slot} & \textbf{Intent} &
			\begin{tabular}[c]{@{}c@{}}\textbf{ATIS} \textbf{Exact}\end{tabular} & \begin{tabular}[c]{@{}c@{}}\textbf{SNIPS} \textbf{Exact}\end{tabular} \\
			\hline
			Full Model & \textbf{84.81} & \textbf{94.30} & \textbf{97.30} & \textbf{86.53} & \textbf{82.62} \\
			\hdashline
			w/o Multi-level Shared-private Architecture & 82.93 & 93.52 & 97.24 & 84.96 & 80.34 \\
			w/o Sentence-level Shared-private Architecture & 82.74 & 93.25 & 96.92 &83.95  & 81.20 \\
			w/o Filter \& Controller & 82.23 & 93.28 & 97.05 & 83.61 & 80.48 \\
			w/o GCN & 83.82 & 93.82 & 96.99 & 85.52 & 81.62 \\
			\hdashline
			w/ Oracle & 87.26 & 94.93 & 96.92 & 86.76 & 87.79 \\
			\hline
		\end{tabular}
	\end{adjustbox}
	\caption{Ablation experiments on ASMixed.}
	\label{tab:ablation}
	\vspace{-0.3cm}
\end{table}

\subsubsection{Effectiveness of Domain-aware Sentence-level Shared-private Framework}
In order to verify the effectiveness of the domain-aware sentence-level shared-private architecture, we conduct the experiment where we adopt the shared context representation $\boldsymbol{c}^{{s}}$ rather than the mixed shared-private $\boldsymbol{c}^{{m}}$ for intent detection and keep other components unchanged. This means that the ablated framework cannot access to domain specific features only for intent detection.
The results are shown in the \textit{w/o Sentence-level Shared-private Architecture} row in Table \ref{tab:ablation}.
We observe that the intent acc drops by 0.38\%, which indicates the sentence-level domain private feature can help intent detection.
In addition, the overall exact drops by 2.07\%. We attribute it to the reason that intent detection and slot filling are the two correlated tasks where the performance of intent detection affects the whole SLU result. 

\subsubsection{Effectiveness of Task-aware Token-Level Shared-Private Framework} \label{sec:task}
To verify the effectiveness of the task-aware token-level shared-private framework, we conduct ablation experiments where we remove the two-stage decoder for slot filling.
In this setting, we incorporate shared and domain-specific features by summation other than using our two-stage slot filling decoder.
This model is effectively a domain-aware version of \texttt{One-Net} \cite{kim2017onenet}, with multi-tasking between intent classification and slot filling only through parameter sharing.
The results are shown in the \textit{w/o Filter \& Controller} row in Table~\ref{tab:ablation}.
We can see a 2.58\% and a 1.02\% drop in the exact and slot filling metrics, respectively, which 
verifies the effectiveness of our proposed two-stage decoder.
We attribute this to the fact that our filter successfully filters the domain-general token and the model automatically learn weights on how to combine shared and domain-specific feature for each token slot prediction.

\subsubsection{Oracle Filter performance}
To see the role of our two-stage decoder intuitively, we also present results when using oracle filter information, by manually filtering out tokens that are not domain-specific. The results are shown in the oracle row.
We obtain 87.26\% on overall exact, outperforming our model over 2.45\%, which demonstrates better two-stage decoder will lead to better multi-domain SLU performance. The result verifies the effectiveness of our two-stage decoder.

\subsubsection{Effectiveness of Domain-aware Syntactic Information}\label{sec:syntactic}
We remove the GCN layers and only adopt the self-attentive encoder to verify the effectiveness of syntax information.
The result is shown in the \textit{w/o GCN} row in Table~\ref{tab:ablation}.
We can see that the performance drops significantly in all metrics, which demonstrates the effectiveness of syntax information in multi-domain SLU tasks.
The reason is that utterances in different domains have different syntactic patterns, which helps the model better capture the shared and domain-specific features.
To our knowledge, we are the first to show that syntactic information is useful for multi-domain SLU.
It is worth noticing that even without the GCN component, our framework still performs the state-of-the-art model \citep{lee-etal-2019-locale}, which again demonstrates the effectiveness and robustness of our framework.

\begin{figure}[t]
	\centering
	\includegraphics[width=0.5\textwidth]{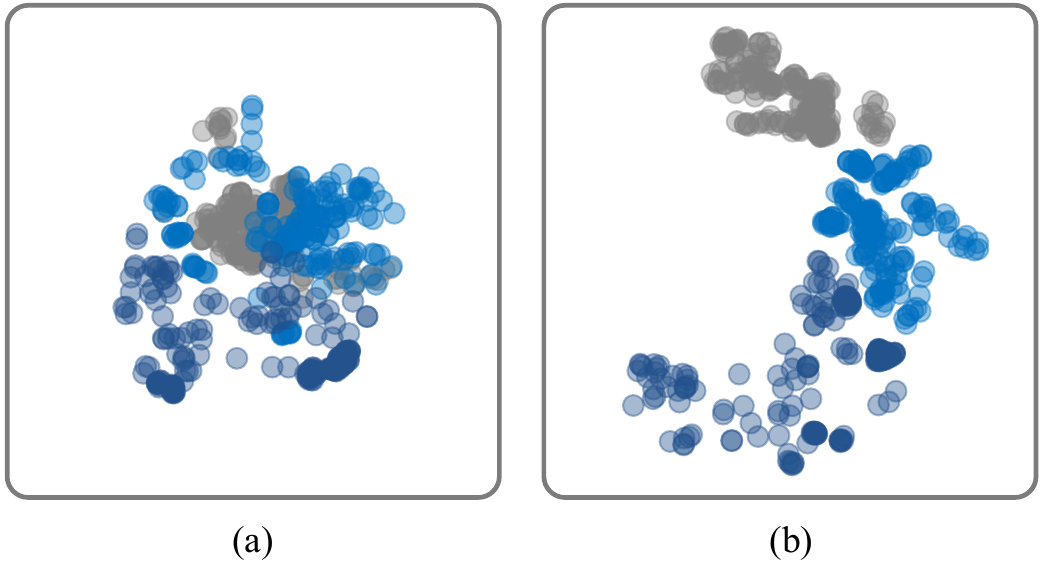}
	\caption{
		t-SNE visualization of sentences vector space from the shared encoder (a) and with each domain private encoder (b). 
	}
	\label{fig:visualize}
\end{figure}

\subsubsection{Domain Adaption} \label{sec:domain_adaption}
We conduct domain adaption experiments  to explore the transferability of our framework on the ASmixed dataset by simulating given a new domain with little labeled data.
We keep ATIS dataset unchanged, and the ratio of the other domain SNIPS from the original data varies from [1\%, 5\%, 10\%, 20\%, 30\%, 40\%, 50\%].
The results are shown in Figure~\ref{fig:few-shot}.
We can find that our framework outperforms \texttt{One-Net} on all ratios of the original dataset.
In particular, our framework trained with 20\% training dataset can achieve comparable and even better performance compared to \texttt{One-Net} with 50\% training dataset on some domains.
In this case, with 5\% training data, our model outperforms \texttt{One-Net} by 12.4\% on SNIPS exact. 
This implies that our framework effectively transfers knowledge from other domains to achieve better performance for the low-resources new domain.

\subsubsection{Breakdown Evaluation} \label{sec:break_down}

In this section, we further investigate why our framework is useful in the few-shot setting, where we keep 5\% original data as training data.
Compared with other domains, we noticed that the slot F1 of the \texttt{Weather} domain outperforms \texttt{One-Net} in the significant range. We conducted a more in-depth analysis of the \texttt{Weather} domain.
As shown in Table \ref{explo}, the slot \texttt{datetime} and \texttt{location} gained the largest improvements.
We observe \texttt{datetime} is a domain-shared slot, which occurs in each domain. It shows the architecture successfully transfers the knowledge among different domains to enhance the performance of the model. 
Moreover, we observe that \texttt{location} is a domain-specific slot that does not have similar slots in other domains, which leads to the difficulty of the prediction of this slot in traditional models.
The results show that the shared-private network structure is useful for improving both domain-shared slots and domain-specific slots by its token-level fine-grained knowledge transfer mechanism.

\begin{table}[t]
	\centering

	\begin{adjustbox}{width=0.7\textwidth}
		\begin{tabular}{l|cccc}
			\hline
			Model & \texttt{location} & \texttt{datetime} & \texttt{noun} &
			\texttt{attribute} \\
			\hline
			One-Net & 81.73 & 90.79 & 94.68 & 91.62 \\
			Our Framework & \textbf{85.46} & \textbf{92.28} & \textbf{95.67} & \textbf{92.24} \\
			\hline
			\multicolumn{1}{c|}{$\Delta$} & +3.73 & +1.49 & +0.99 & +0.62\\
			\hline
		\end{tabular}
	\end{adjustbox}
	\caption{The delta of major slots' F1 in \texttt{Weather} domain with 5\% training data.}
	\label{explo}
	\vspace{-0.3cm}
\end{table}

\subsubsection{Visualization} \label{sec:visual}
To understand whether our framework successfully captures the domain-shared and domain-specific features, we visualize $\boldsymbol{G}^s$ and $\boldsymbol{G}^p$ from the full model trained on MTOD. In particular, we put 600 sentences which include 200 sentences from the Alarm domain, 200 sentences from the Reminder domain, and 200 sentences from \texttt{Weather} domain into the shared syntax-aware encoder.
Sentences from each domain are fed into its private syntax-aware encoder to get their sentence representations. 

We use t-SNE to visualize sentence representations obtained by the shared and private encoders.
The vectors are shown in Figure~\ref{fig:visualize}.
We can observe that those representations from the shared encoder tend to stay closer.
In contrast, each private sentence representations from each domain tend to occur in a cluster, and there is nearly no overlap between different domains. This demonstrates our syntax-aware encoders capture the domain-shared and domain-specific features effectively.

\begin{figure}[t]
	\centering
	\includegraphics[width=0.7\textwidth]{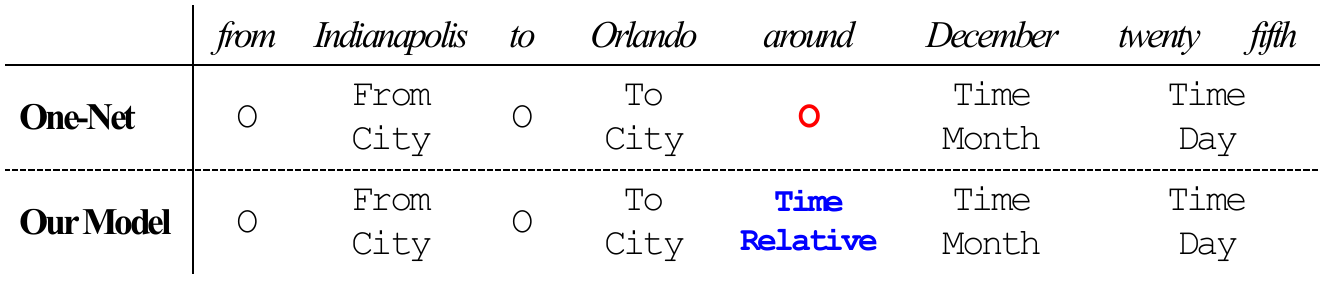}
	\caption{
		Case Study.
		The \textcolor{blue} {blue slot} is correct while the  \textcolor{red} {red one} is wrong.
		Better viewed in color.
	}
	\label{fig:case}\end{figure}

\subsubsection{Case Study} \label{sec:case}
To better understand how our proposed task-aware two-stage decoder affects and contributes to the final result,
we conduct a
case study of the slot filling task between our model and the baseline model \texttt{One-Net}.

This case is shown in Figure~\ref{fig:case}.
For the word ``\textit{around}'', \texttt{One-Net} predicts its slot label as ``\texttt{O}'' incorrectly.
The token``\textit{around}'' is more likely treated as a domain-general token because it usually does not have real meaning in SLU system.
The result indicates that \texttt{One-Net} cannot capture sufficient domain information to predict it correctly. 
In contrast, our model predicts the slot label correctly.
We attribute this to the fact that the proposed  two-stage decoder successfully learns to capture more 
domain-specific knowledge for this token, achieving the fine-grained knowledge transfer.
\begin{figure}[t]
	\centering
	\includegraphics[width=0.7\textwidth]{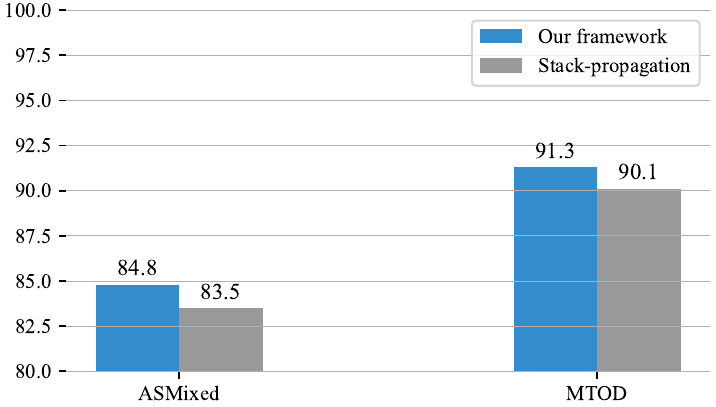}
	\caption{
		Performance (Exact Acc) on two datasets between our framework with the SOTA single-domain model (Stack-propagation).
	}
	\label{fig:stack}
\end{figure}
\subsubsection{Compared with the Best Single-domain Model}
To further verify the effectiveness of our proposed method, we compare our model with the state-of-the-art single-domain model \texttt{Stack-propagation} \citep{qin-etal-2019-stack}.
\texttt{Stack-propagation} is directly trained with the mixed dataset, which can be considered as a shared model to implicitly extract domain knowledge.

The results are shown in Figure~\ref{fig:stack}, we can observe that our framework outperforms Stack-propagation on two datasets, which demonstrates that our proposed multi-level shared-private framework makes better domain knowledge representation and transfer than the single-domain-based model.

\subsubsection{Effect of RoBERTa}
A natural question raised that the pre-trained models (PLMs) have achieved surprising results across almost all NLP tasks, whether our framework can still obtain improvement over the pre-trained model.
To answer the question, we explore the RoBERTa \citep{liu2019roberta} on our framework and we name it as \texttt{our framework + RoBERTa}. 
More specifically, we replace the shared Syntactic Encoder with RoBERTa-base and keep other components unchanged. 
In the experimental setting, we adopt the fine-tuning mode.
For generating token hidden representation, we follow \citet{qin-etal-2019-stack} to consider the first subword label if a word is broken into multiple subwords. 
For example, if sentence ``\textit{\texttt{[<s>]} The movie is very interesting \texttt{[</s>]}}'' is split into ``\textit{\texttt{[<s>]} The movie is very inter\#\# \#\#esting \texttt{[</s>]}}'', we only adopt the representation of \textit{inter} as the whole token \textit{interesting} representation.

\begin{figure}[t]
	\centering
	\includegraphics[width=0.7\textwidth]{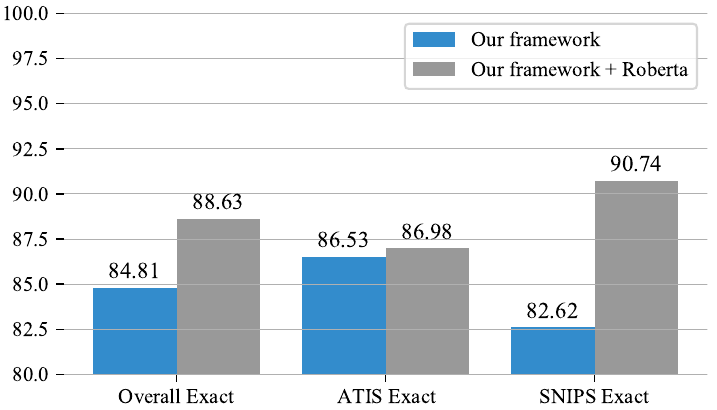}
	\caption{
		Performance (Exact Acc) on two datasets between our framework and our framework+RoBERTa.
	}
	\label{sec:exp:all_data_roberta}
\end{figure}

The comparison results are shown in Figure~\ref{sec:exp:all_data_roberta}, we find that our \texttt{framework + RoBERTa} outperforms our model on all datasets, which indicates the effectiveness of the pre-trained model.
We attribute this to the fact that pre-trained models can provide rich semantic features, which can help to improve the performance on multi-domain SLU tasks, which has the consistent observation with \citet{qin2020dcr}.

\section{Related Work}
\paragraph{Intent Detection and Slot Filling}
Intent detection and slot filling are two core subtasks of spoken language understanding (SLU), which aims to identify users’ intents and to extract semantic constituents from the natural language utterances \citep{tur2011spoken}.
Intent detection can be considered as the sentence classification task.
The classical methods such as support vector machine (SVM) \cite{haffner2003optimizing} and RNN\cite{sarikaya2011deep}, have been proposed to solve intent detection. 
Recently, \citet{xia-etal-2018-zero} adopts a capsule-based neural network with self-attention for intent detection, achieving the promising performance.

Slot filling can be regarded as  a sequence labeling task.
The popular approaches are conditional random fields (CRF) \citep{raymond2007generative} and recurrent neural networks (RNN) \citep{xu2013convolutional,yao2014spoken}. 
Recently, \citet{qin-etal-2019-stack}, \citet{shen2018disan}, \citet{tan2018deep}  and \citet{qin2020co} propose the self-attention mechanism sequential labeling, which achieves the promising performance without CRF structure.

\paragraph{Joint Model for SLU}
Since intent and slots are closely related, dominant methods \citep{zhang2016joint, goo-etal-2018-slot, li-etal-2018-self, wang2018bi,qin-etal-2019-stack,zhang-etal-2019-joint,teng2021injecting,qin2020cosdaml,qin2021survey} in the literature adopt the joint model to consider the mutual relationship between slot filling and intent detection.
\citet{zhang2016joint} proposed a joint model using LSTM for learning the correlation between intent and slots.
\citet{goo-etal-2018-slot} proposed a slot-gated model to consider the relationship and interaction between two tasks.
\citet{li-etal-2018-self} and \citet{qin-etal-2019-stack} proposed to use intent information to explicitly model the semantic correlation between slots and intent.
However, the above work is restricted to a single domain. In contrast, we consider joint SLU in a multi-domain setting.

\paragraph{Multi-Domain SLU}
\citet{hakkani2016multi} proposed a single LSTM model over a mixed multi-domain dataset implicitly learning the domain-shared features.
\citet{kim2017onenet} adopted one network to jointly modeling slot filling, intent detection and domain classification.
\citet{liu2020coach} proposed a coarse-to-fine approach (Coach)
for cross-domain slot filling.
The above methods trained a single model on the mixed dataset.
Compared with their work, we propose a domain-aware and task-aware model by extending a shared-private framework into a token-level knowledge sharing structure.
In addition, the above works do not incorporate syntax information, we find that modeling the syntax information is useful for multi-domain SLU.
More closely related to our work, \citet{lee-etal-2019-locale} used a shared-private framework for domain classification, and \citet{liu2017multi} used a shared-private framework for slot filling. These methods can be regarded as application of the standard shared-private architecture to subproblems in SLU. In contrast to their work, we exploit the mutual benefit between intent detection and slot filling for fine-grained knowledge transfer. To our knowledge, we are the first to investigate shared-private framework for \textit{joint} SLU, and the first to conduct the token-level selective weighing shared and private representations in their integration.
\paragraph{Graph Convolutional Network}
Graph convolutional networks (GCN) are neural networks that operate directly on graph structures \citep{kipf2016semi} to model the structural information, which has been applied successfully in various NLP tasks. 
\citet{de-cao-etal-2019-question} and \citet{lin-etal-2019-kagnet} propose GCN to perform multi-step reasoning on question answering task.
\citet{strubell2018linguistically} and \citet{marcheggiani-titov-2017-encoding} utilize GCN to model the syntactic information for semantic role labeling. \citet{huang-carley-2019-syntax} and \citet{zhang-etal-2019-aspect} apply GCN for aspect-based sentiment classification to consider syntactical constraints.
\citet{qin-etal-2020-agif,qin-etal-2021-gl} explore the graph network to model the interaction between the slot and multiple intents.
\citet{guo-etal-2019-attention,guo-etal-2019-densely} successfully propose an attention guided graph convolutional network to encode the dependency trees for relation extraction. 
Our work follows the above line of models. We propose to utilize the GCN to explore the graph structure to encode the syntactic information for multi-domain SLU tasks.
To the best of our knowledge, we are the first to incorporate syntactic information with GCN for multi-domain SLU.

\section{Conclusion}
We investigated domain-aware and task-aware parameterization for multi-domain SLU by building
a model with separate domain- and task-specific
parameters. 
In particular, a domain-aware sentence-level shared-private framework can be used for extracting domain knowledge for intent detection while a task-aware token-level shared-private framework is used to achieve a fine-grained knowledge transfer for slot filling.
Unlike existing methods, which use the same parameters for multi-task learning, our model can achieve a fine-grained combination of domain knowledge transfer.
Experiments on two publicly available datasets with five domains show the effectiveness of the proposed models and we achieve the state-of-the-art performance.
In addition, our model can quickly adapt to a new domain given little labeled data, which makes it more robust and scalable in the real-world scenario.
Finally, our work is the first attempt to explore syntax information and empirically demonstrate the effectiveness of syntax information in multi-domain SLU tasks.

\section*{Acknowledgment}
This work was supported by Westlake-BrightDreams Robotics research grant.
Besides, this work was also supported by the National Key R\&D Program of China via grant 2020AAA0106501 and the National Natural Science Foundation of China (NSFC) via grant 61976072 and 61772153.
This work was also supported by the Zhejiang Lab’s International Talent Fund for Young Professionals.

\bibliographystyle{ACM-Reference-Format}
\bibliography{taslp}

\end{document}